\documentclass[a4paper,10pt,fleqn]{article}

\usepackage[a4paper, margin=1in]{geometry}
\usepackage{graphicx}
\usepackage{fancyhdr}
\usepackage{titlesec}
\usepackage{setspace}
\usepackage{amsmath,amssymb}
\usepackage{enumitem}
\usepackage{hyperref}
\usepackage{totpages}
\usepackage{caption}
\usepackage{float}
\usepackage{booktabs}
\usepackage{algorithm}
\usepackage{algpseudocode}
\usepackage{tikz}
\usetikzlibrary{positioning,arrows.meta,shapes.geometric,fit,backgrounds}
\usepackage{IAAConference}

\begin{document}

\newcommand{\customauthor}[3]{\textbf{#1}\textsuperscript{#2}\ifnum#3=1\textsuperscript{*}\fi}
\newcommand{\customaffiliation}[2]{\textsuperscript{#1}\textit{#2}}

\newcommand{\authorslist}{
    \customauthor{May Hammad}{1}{1},
    \customauthor{Menatallh Hammad}{1}{1},
}

\newcommand{\affiliationslist}{
    \customaffiliation{1}{Department of Aerospace Informatics, Julius-Maximilians-Universit\"at W\"urzburg\\W\"urzburg, Germany}
}

\begin{center}
    \textbf{Paper number authorized by the organization committee}\\[1.5em]
    \textbf{\large Star-Fusion: Multi-modal Transformer Architecture for Coarse Celestial Boresight Localization}\\[1.5em]
    \authorslist\\[0em]
    \affiliationslist\\[0em]
    \textit{* Corresponding Author}\\[2em]
\end{center}

\renewenvironment{abstract}{
    \begin{center}
        \textbf{Abstract}
    \end{center}
    \noindent\begin{spacing}{1}
}{
    \end{spacing}\vspace{1em}
}

\begin{abstract}
Reliable celestial localization is an important component of autonomous spacecraft navigation, particularly during Lost-in-Space (LIS) acquisition when no prior attitude estimate is available. Classical star-identification pipelines are accurate and geometrically well founded, but they can incur nontrivial catalog-matching overhead and may degrade when detections are corrupted by sensor noise, missing stars, or false observations. Learning-based alternatives offer a complementary route, but direct regression of celestial angular coordinates is complicated by the periodicity of Right Ascension (RA) and by coordinate singularities near the celestial poles.

This paper presents Star-Fusion, a multi-representation architecture for \emph{coarse celestial boresight localization}. Rather than regressing RA and Declination (Dec) directly, the method maps each boresight direction to a unit vector on the sphere and assigns it to one of $K$ spherical clusters. The network then predicts the corresponding celestial region from three complementary representations of the same observation: a SwinV2 image encoder for photometric information, a CNN operating on a rasterized star heatmap, and an MLP operating on detected-star coordinates. On a synthetic Stellarium -derived dataset with $K=15$, Star-Fusion achieves 93.4\% Top-1 accuracy and 97.8\% Top-3 accuracy on the held-out validation set, compared with 84.2\% Top-1 accuracy for a SwinV2-Tiny image-only baseline. The neural forward pass requires 18.4~ms on an embedded NVIDIA Jetson platform operated in a 15~W power mode. These results support Star-Fusion as a coarse LIS localization module, while full three-axis attitude recovery and validation under real sensor conditions remain outside the scope of the present study.
\end{abstract}

\noindent \textbf{Keywords:} Celestial Localization, Star Tracker, Lost-in-Space, Multi-modal Transformer, Spherical K-Means, Embedded Spacecraft Navigation.

\section{Introduction}
Spacecraft attitude determination estimates the orientation of a vehicle relative to an inertial reference frame. Star trackers are among the most accurate sensors for this task because distant stellar constellations provide an effectively fixed celestial reference. Classical Lost-in-Space (LIS) pipelines typically perform star centroiding, star identification against an onboard catalog, and subsequent attitude recovery. Algorithms such as Pyramid~\cite{mortari1997pyramid} and Grid~\cite{padgett1997grid} are mathematically rigorous and widely studied, but catalog matching and geometric verification can become computationally demanding in difficult observation conditions, including sensor noise and false detections.

Learning-based alternatives have therefore received increasing attention. Rijlaarsdam et al.~\cite{rijlaarsdam2020efficient} demonstrated that a compact fully connected network can replace database lookups for star identification, while Wang et al.~\cite{wang2020ai} showed that a VGG16-based classifier can retain strong identification performance under position noise, magnitude noise, and false-star interference. Wang et al.~\cite{wang2021cnn} further proposed a rotation-invariant 1D-CNN operating on log-polar star patterns to improve robustness to distorted observations. These studies show that neural models can provide useful learned priors for star-pattern recognition.

The scope of the present work is deliberately narrower than full three-axis attitude determination. Star-Fusion predicts a \emph{coarse celestial boresight region}, represented by RA and Dec, and therefore estimates two angular degrees of freedom rather than the complete spacecraft orientation in $SO(3)$. The intended use is as a coarse LIS acquisition stage that narrows the celestial search space before a downstream fine attitude solver.

\subsection{Motivation and problem statement}
A direct regression target expressed in RA and Dec has an inconvenient coordinate representation. Right Ascension is periodic, so directions near $0^\circ$ and $360^\circ$ are physically adjacent despite being numerically distant. Declination is not periodic, but the usual RA/Dec parameterization becomes ill-conditioned near the poles because changes in RA correspond to vanishing geodesic motion as $|\delta|\rightarrow 90^\circ$. A Euclidean loss in the raw angular coordinates therefore does not faithfully represent angular proximity on the celestial sphere.
\\
Star-Fusion addresses this issue by first mapping each ground-truth boresight direction onto the unit sphere and then converting the continuous direction into a discrete spherical region. This removes the RA wrap-around discontinuity from the learning target and makes the output naturally interpretable as a coarse celestial hypothesis. The model then combines photometric and geometric evidence through three encoders operating on different representations of the same star-field observation.

\subsection{Contributions}
The main contributions are:
\begin{itemize}
    \item \textbf{Spherical discretization for coarse LIS localization:} RA/Dec labels are mapped to 3D unit vectors and partitioned using a normalized spherical K-Means procedure, avoiding the artificial discontinuity of raw RA regression.
    \item \textbf{Three-branch multi-representation fusion:} The model combines a SwinV2 image encoder with a CNN over a rasterized star heatmap and an MLP over detected-star coordinates. The heatmap and coordinate branches encode the same geometric detections through different inductive biases rather than independent information sources.
    \item \textbf{Embedded forward-pass evaluation:} The neural inference stage is evaluated on an embedded NVIDIA Jetson platform under a 15~W device power configuration, providing an initial measure of onboard computational feasibility.
\end{itemize}

\section{Mathematical Formulation}
\subsection{Spherical topology mapping}
Let $x\in\mathbb{R}^{H\times W}$ denote a star-field observation and let the ground-truth boresight direction in the ICRS frame aka. (RA \& Dec), be $\Theta=(\alpha,\delta)$, where $\alpha$ is Right Ascension and $\delta$ is Declination. The direction is embedded on the unit sphere through
\begin{equation}\label{eq:spherical_mapping}
\phi(\alpha,\delta)=
\begin{bmatrix}
\cos\delta\cos\alpha & \cos\delta\sin\alpha & \sin\delta
\end{bmatrix}^{T}.
\end{equation}
The resulting vector satisfies $\|\phi(\alpha,\delta)\|_2=1$.

The discrete class label is obtained from $K$ unit-norm centroids $C=\{\mathbf{c}_1,\ldots,\mathbf{c}_K\}$ using maximum cosine similarity,
\begin{equation}\label{eq:cluster_label}
    y=\arg\max_{k\in\{1,\ldots,K\}} \phi(\Theta)^T\mathbf{c}_k.
\end{equation}
For unit vectors, this assignment is equivalent to minimizing Euclidean distance on the embedded sphere. Centroids are initialized using K-Means++ and re-normalized after each update.

\begin{algorithm}[H]
\caption{Spherical pseudo-label generation}
\label{alg:spherical_labels}
\begin{algorithmic}[1]
\Require Training directions $\{(\alpha_i,\delta_i)\}_{i=1}^{N}$
\Ensure Unit-norm centroids $C$ and cluster labels $Y$
\For{each training sample $i$}
    \State $\mathbf{v}_i\gets[\cos\delta_i\cos\alpha_i,\cos\delta_i\sin\alpha_i,\sin\delta_i]^T$
\EndFor
\State Initialize $K$ centroids with K-Means++
\State Normalize each centroid: $\mathbf{c}_k\gets\mathbf{c}_k/\|\mathbf{c}_k\|_2$
\While{not converged}
    \State Assign $y_i\gets\arg\max_k\mathbf{v}_i^T\mathbf{c}_k$
    \For{$k=1,\ldots,K$}
        \State $\tilde{\mathbf{c}}_k\gets\sum_{i:y_i=k}\mathbf{v}_i$
        \State $\mathbf{c}_k\gets\tilde{\mathbf{c}}_k/\|\tilde{\mathbf{c}}_k\|_2$
    \EndFor
\EndWhile
\State \Return $C,Y$
\end{algorithmic}
\end{algorithm}

For validation or test observations, the training-derived centroids are kept fixed and labels are assigned using Eq.~\eqref{eq:cluster_label}. This avoids defining the class partition from held-out data.

\section{Star-Fusion Architecture}
The Star-Fusion model contains three parallel encoders, $E_p$, $E_g$, and $E_c$, representing the photometric image, rasterized geometry, and coordinate representation, respectively. Figure~\ref{fig:architecture} summarizes the end-to-end pipeline.

\begin{figure}[H]
\centering
\begin{tikzpicture}[
    font=\scriptsize,
    node distance=5mm and 6mm,
    input/.style={rectangle, rounded corners, draw=black!70, fill=black!5, minimum width=2.1cm, minimum height=0.8cm, align=center},
    enc/.style={rectangle, rounded corners, draw=black!70, fill=blue!8, minimum width=2.1cm, minimum height=0.9cm, align=center},
    feat/.style={rectangle, draw=black!60, fill=white, minimum width=1.9cm, minimum height=0.7cm, align=center},
    fuse/.style={rectangle, rounded corners, draw=black!70, fill=orange!12, minimum width=2.6cm, minimum height=0.8cm, align=center},
    outnode/.style={rectangle, rounded corners, draw=black!70, fill=green!10, minimum width=2.6cm, minimum height=0.8cm, align=center},
    arr/.style={-{Latex[length=1.8mm]}, thick}
]
    \node[input] (image) {Image $I$};
    \node[enc, right=of image] (swin) {SwinV2-Tiny $E_p$};
    \node[feat, right=of swin] (hp) {$\mathbf{h}_p\in\mathbb{R}^{768}$};

    \node[input, below=of image] (heatmap) {Heatmap $H$};
    \node[enc, right=of heatmap] (cnn) {Shallow CNN $E_g$};
    \node[feat, right=of cnn] (hg) {$\mathbf{h}_g$};

    \node[input, below=of heatmap] (coords) {Coordinates $s$};
    \node[enc, right=of coords] (mlp) {Coordinate MLP $E_c$};
    \node[feat, right=of mlp] (hc) {$\mathbf{h}_c$};

    \node[fuse, right=13mm of hg] (concat) {Concatenate\\$[\mathbf{h}_p\parallel\mathbf{h}_g\parallel\mathbf{h}_c]$};
    \node[fuse, below=5mm of concat] (ln) {LayerNorm};
    \node[outnode, below=5mm of ln] (linear) {Linear + Softmax};
    \node[input, below=5mm of linear] (pred) {$P(y\mid I,H,s)$};

    \draw[arr] (image) -- (swin);
    \draw[arr] (heatmap) -- (cnn);
    \draw[arr] (coords) -- (mlp);
    \draw[arr] (swin) -- (hp);
    \draw[arr] (cnn) -- (hg);
    \draw[arr] (mlp) -- (hc);

    \draw[arr] (hp) -| (concat.north);
    \draw[arr] (hg) -- (concat);
    \draw[arr] (hc) -| (concat.south);

    \draw[arr] (concat) -- (ln);
    \draw[arr] (ln) -- (linear);
    \draw[arr] (linear) -- (pred);

    \begin{scope}[on background layer]
        \node[draw=black!40, dashed, rounded corners, fit=(image)(heatmap)(coords)(swin)(cnn)(mlp)(hp)(hg)(hc), inner sep=5pt, label={[font=\scriptsize]above:Three-branch encoders}] {};
    \end{scope}
\end{tikzpicture}
\caption{Star-Fusion pipeline. The photometric branch processes the rendered sensor image, while the heatmap and coordinate branches encode two representations of the detected star geometry. Their latent features are concatenated, normalized, and mapped to a categorical distribution over $K$ spherical boresight regions.}
\label{fig:architecture}
\end{figure}

\subsection{Photometric encoder}
The photometric encoder uses SwinV2-Tiny~\cite{liu2022swinv2} to extract a feature vector $\mathbf{h}_p\in\mathbb{R}^{768}$. The shifted-window mechanism permits information exchange across local attention windows and provides a large effective receptive field for star-pattern recognition. This choice builds on earlier evidence that convolutional backbones such as VGG16 can provide strong robustness for learned star identification~\cite{wang2020ai}.

\subsection{Geometric and coordinate encoders}
Let $s$ denote the detected-star coordinate representation supplied to the geometric subsystem. The heatmap $H$ is obtained by rasterizing these detections as Gaussian kernels on a fixed canvas, so $H$ is a deterministic representation of the same underlying geometry rather than an independent sensor modality. A shallow CNN maps $H$ to a latent feature $\mathbf{h}_g=E_g(H)$, while a three-layer MLP maps $s$ to $\mathbf{h}_c=E_c(s)$.


Because a raw star set has no intrinsic ordering, the exact ordering and padding convention used to construct the MLP input must be stated explicitly. If the implementation does not enforce a deterministic ordering, a permutation-invariant set encoder would be a more principled replacement in future work.

\subsection{Fusion logic}
For an input triplet $(I,H,s)$, the branch features are concatenated and normalized,
\begin{equation}\label{eq:feature_integration}
\mathbf{h}_{\mathrm{fusion}}=
\mathrm{LayerNorm}([E_p(I)\parallel E_g(H)\parallel E_c(s)]).
\end{equation}
The class logits are then
\begin{equation}\label{eq:logits}
\boldsymbol{\ell}=W_o\mathbf{h}_{\mathrm{fusion}}+\mathbf{b}_o,
\end{equation}
and the output distribution is
\begin{equation}\label{eq:softmax}
P(y=k\mid I,H,s)=\frac{\exp(\ell_k)}{\sum_{j=1}^{K}\exp(\ell_j)}.
\end{equation}
Layer normalization is used to stabilize the scale of the fused representation before classification; it is not assumed to guarantee equal contribution from the three branches.

\begin{algorithm}[H]
\caption{Star-Fusion forward pass}
\label{alg:forward_pass}
\begin{algorithmic}[1]
\Require Image $I$, heatmap $H$, star coordinates $s$
\Ensure Class probabilities $P$
\State $\mathbf{f}_{\mathrm{swin}}\gets\mathrm{SwinV2}(I)$
\State $\mathbf{h}_p\gets\mathrm{GlobalAveragePool}(\mathbf{f}_{\mathrm{swin}})$
\State $\mathbf{h}_g\gets\mathrm{Flatten}(\mathrm{CNN}(H))$
\State $\mathbf{h}_c\gets\mathrm{MLP}_{\mathrm{coords}}(s)$
\State $\mathbf{z}\gets[\mathbf{h}_p\parallel\mathbf{h}_g\parallel\mathbf{h}_c]$
\State $\mathbf{z}_{\mathrm{norm}}\gets\mathrm{LayerNorm}(\mathbf{z})$
\State $\mathrm{logits}\gets\mathrm{Linear}(\mathbf{z}_{\mathrm{norm}})$
\State $P\gets\mathrm{Softmax}(\mathrm{logits})$
\State \Return $P$
\end{algorithmic}
\end{algorithm}

\subsection{Heatmap representation}
Each detected star is rendered as a Gaussian kernel on a fixed heatmap canvas, converting a sparse point set into a dense spatial representation suitable for convolutional processing. The number of visible stars per sample varies with field of view and limiting magnitude and is allowed to range between 2 and 6, while the heatmap dimensions and Gaussian bandwidth remain fixed during training.


\begin{figure}[H]
\centering
\includegraphics[width=0.95\textwidth]{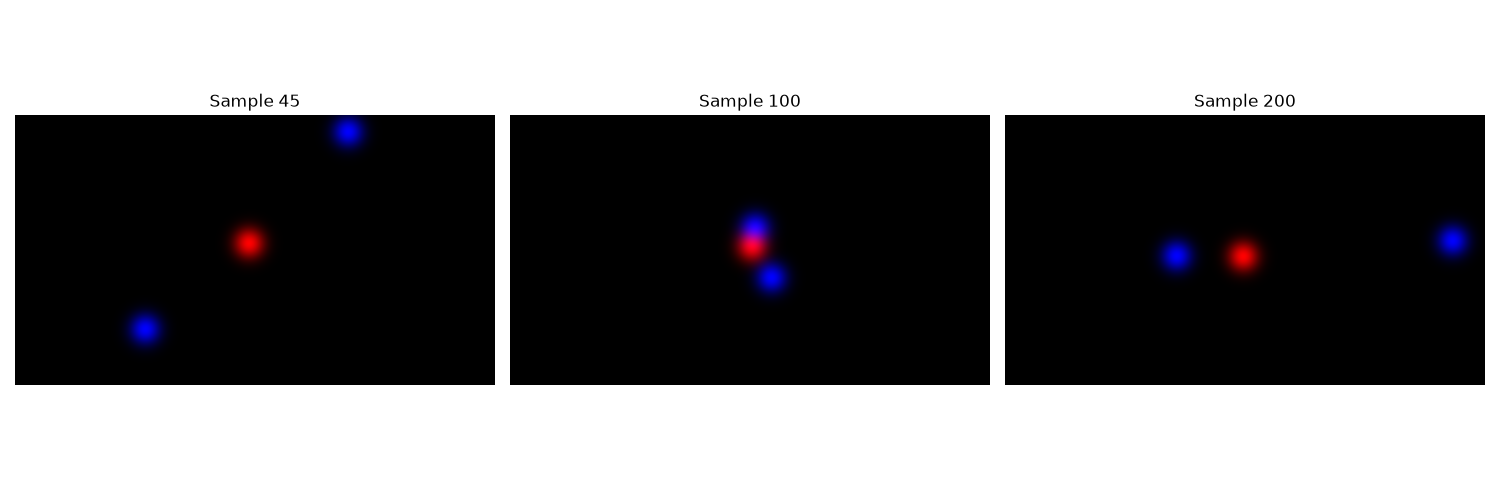}
\caption{Representative rasterized star heatmaps for three validation samples. Colors in this figure should be interpreted only according to the actual network input implementation; any visualization-only highlighting must not be provided as privileged target information to the model. The heatmap and coordinate branches encode different representations of the same detected star geometry.}
\label{fig:heatmaps}
\end{figure}

\section{Experiments and Results}
\subsection{Dataset and training configuration}
The model was evaluated on a synthetic dataset of 50,000 star-field images generated from the Stellarium Telescope collected dataset, with 40,000 images used for training and 10,000 images used as a held-out validation set. Images were rendered at $1024\times1024$ resolution and downsampled to $256\times256$ for the photometric branch. The classification objective was a focal loss,
\begin{equation}
    \text{FL} = -\sum_{c=1}^{C} \alpha_c y_c (1 - p_c)^\gamma \log(p_c)
    \label{eq:focal_loss_multiclass}
\end{equation}
where $C=K=15$ is the number of spherical classes, $y_c\in\{0,1\}$ is the
one-hot ground-truth indicator for class $c$, and $p_c\in[0,1]$ is the
predicted softmax probability for class $c$. The focusing parameter was set
to $\gamma=2.0$. The per-class weight $\alpha_c$ was computed using the
standard inverse-class-frequency (``balanced'') heuristic,
$\alpha_c \propto N/(K\,n_c)$, where $N$ is the total number of training
samples and $n_c$ is the number of samples in class $c$, followed by
$L_1$ normalization so that $\sum_{c=1}^{K}\alpha_c=1$.
\\
The network was trained with AdamW ($\beta_1=0.9$, $\beta_2=0.999$, weight
decay $0.05$), an initial learning rate of $3\times10^{-4}$ decayed on a
cosine schedule to $1\times10^{-6}$, with a 5-epoch linear warmup. Training
used a batch size of 64. The SwinV2-Tiny branch was initialized from
weights pretrained on ImageNet-1K at $256\times256$ resolution
(\texttt{microsoft/swinv2-tiny-patch4-window8-256}), while the heatmap CNN
and coordinate MLP branches were trained from scratch. Data augmentation
included random in-plane rotation ($\pm 5$--$10^{\circ}$) around the
reference star, brightness/contrast jitter, additive Gaussian pixel noise,
random star dropout to simulate missed detections, and random false-star
injection. Training was configured for a maximum of 80 epochs, with early
stopping (patience of 10 epochs on the validation loss) triggered once
performance ceased to improve; the best-performing checkpoint, at epoch
64, was retained for evaluation.
\\
Training converged at epoch 64, with a final training loss of 0.011 and
validation loss of 0.072.
\\
The present study reports validation performance from a single training
run, because no independent test split was used and results were not
averaged over multiple random seeds. This should be interpreted as
preliminary empirical evidence rather than a final estimate of deployment
performance or its variance.
\\

\begin{figure}[H]
\centering
\includegraphics[width=0.75\textwidth]{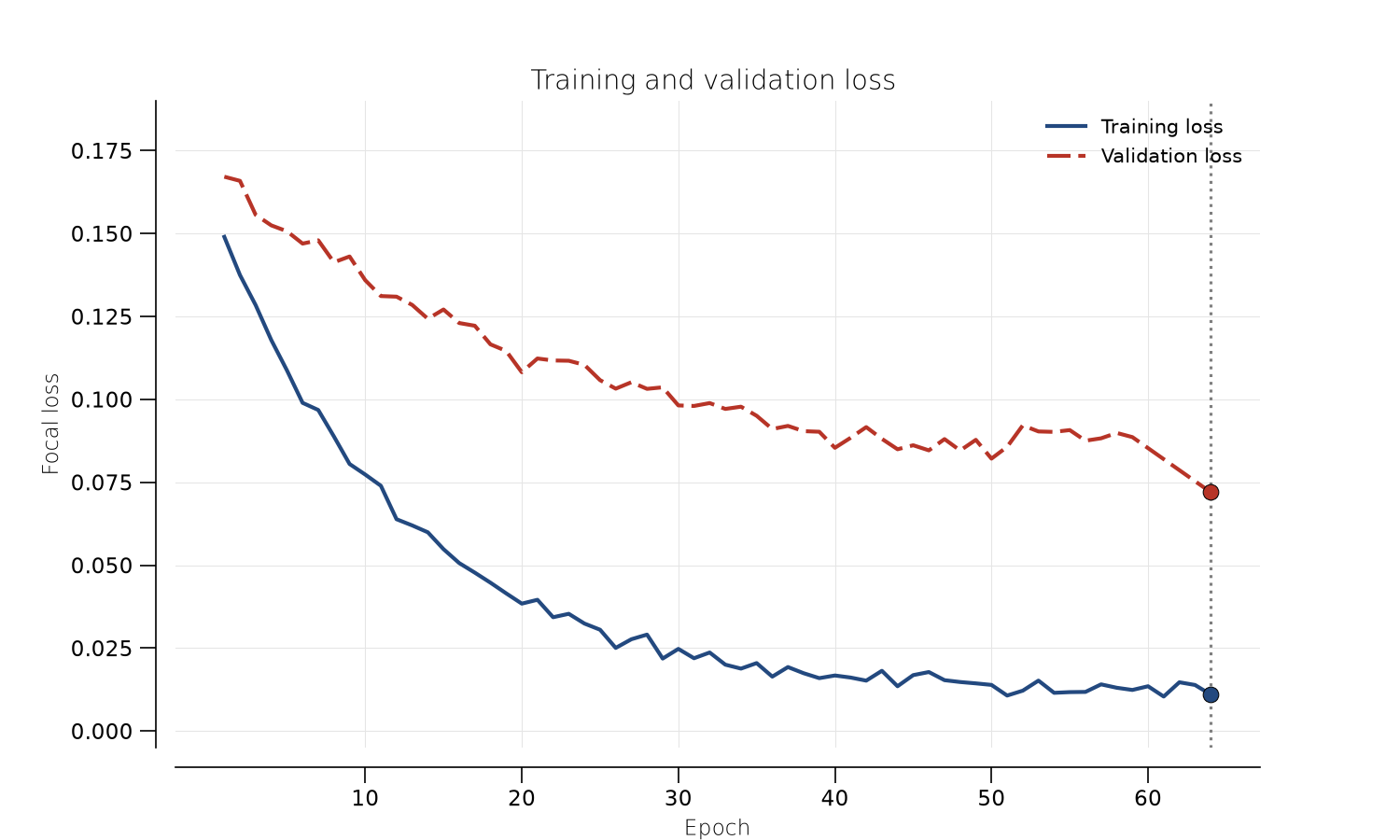}
\caption{\textbf{Training dynamics and edge deployment performance of the
proposed star tracker model.} (a) Progression of training and validation
focal loss over 64 epochs. (b) The applied learning rate schedule,
consisting of a 5-epoch linear warmup followed by cosine annealing decay
from $3\times10^{-4}$ to $1\times10^{-6}$. (c) Single-frame inference
latency (batch size~1) on an NVIDIA Jetson Nano using FP16 precision, with
a measured mean processing time of 18.4~ms, satisfying real-time processing
requirements.}
\label{fig:loss_curve}
\end{figure}

\subsection{Classification performance}
Star-Fusion was compared against an image-only SwinV2-Tiny baseline and a ResNet-50 baseline. Table~\ref{tab:performance} reports Top-1, Top-3, and Top-5 classification accuracy over the $K=15$ coarse spherical regions.

\begin{table}[H]
\centering
\caption{Validation performance of Star-Fusion and image-only baselines for $K=15$ spherical regions.}
\label{tab:performance}
\begin{tabular}{lccc}
\toprule
Model & Top-1 Acc. & Top-3 Acc. & Top-5 Acc. \\
\midrule
ResNet-50 & 79.4\% & 88.2\% & 92.1\% \\
SwinV2-T Baseline~\cite{liu2022swinv2} & 84.2\% & 91.5\% & 95.8\% \\
Star-Fusion & 93.4\% & 97.8\% & 99.2\% \\
\bottomrule
\end{tabular}
\end{table}

Star-Fusion improves Top-1 accuracy by 9.2 percentage points over the SwinV2-Tiny baseline. Because the task contains only 15 coarse classes, Top-3 and especially Top-5 accuracy should be interpreted as candidate-region recall rather than fine angular precision. In a practical LIS pipeline, multiple high-probability regions could be passed to a downstream geometric verifier or fine attitude estimator.

A classification result does not directly quantify boresight angular error. For a predicted class $\hat{y}$ with centroid $\mathbf{c}_{\hat{y}}$, a physically interpretable coarse localization error can be defined as
\begin{equation}\label{eq:angular_error}
e_{\mathrm{ang}}=
\arccos\left(\phi(\Theta)^T\mathbf{c}_{\hat{y}}\right).
\end{equation}

\subsection{Ablation study}
An ablation study was used to estimate the marginal utility of the geometric representations. Removing the coordinate branch reduced Top-1 accuracy from 93.4\% to 88.9\%, a decrease of 4.5 percentage points, while removing the heatmap branch reduced accuracy to 90.1\%.

\begin{table}[H]
\centering
\caption{Ablation study on feature branches. Changes are reported in percentage points relative to the full model.}
\label{tab:ablation}
\begin{tabular}{lc}
\toprule
Configuration & Top-1 accuracy \\
\midrule
Full Star-Fusion & 93.4\% \\
Without heatmap branch & 90.1\% ($-3.3$ pp) \\
Without coordinate branch & 88.9\% ($-4.5$ pp) \\
Unimodal SwinV2 only & 84.2\% ($-9.2$ pp) \\
\bottomrule
\end{tabular}
\end{table}

These results show that both geometric representations contribute beyond the image-only baseline. However, the present ablation does not isolate the absolute strength of each representation. In particular, coordinate-only, heatmap-only, image-plus-coordinate, image-plus-heatmap, and coordinate-plus-heatmap variants would be required to distinguish true fusion gains from a dominant individual branch.

\subsection{Computational complexity and onboard feasibility}
For power-constrained spacecraft systems, inference cost is an important design consideration. The Star-Fusion network contains 29.8~M parameters and requires 4.6~GFLOPs per forward pass. On an embedded NVIDIA Jetson platform configured in a 10~W device power mode, the measured neural inference latency was 18.4~ms, corresponding to 54.3~frames per second.
\\
This measurement demonstrates that the \emph{neural forward pass} can execute at real-time rates on embedded hardware. It should not be interpreted as complete end-to-end system latency unless preprocessing, centroid extraction, coordinate construction, heatmap rendering, memory transfer, and downstream attitude refinement are included in the timing budget. Likewise, the 10~W setting refers to the platform power configuration and is not a direct measurement of network power consumption.

\subsection{Error analysis}
The reported confusion analysis indicates that 62\% of misclassifications occur between geographically neighboring clusters near the celestial equator, suggesting that a substantial fraction of errors are associated with coarse quantization boundaries rather than arbitrary jumps across the sphere.


This observation motivates geometry-aware targets in future work. Instead of assigning all probability mass to a single Voronoi cell, nearby centroids could receive soft target weights as a function of spherical distance.

\section{Discussion}
\subsection{Complementary representations of star geometry}
The photometric branch captures appearance information from the rendered star field, including relative brightness and broader image context. The heatmap branch removes much of this photometric variation and instead emphasizes the spatial arrangement of detected stars. The coordinate MLP provides a direct numerical representation of the same detections. The latter two branches are therefore not independent modalities in an information-theoretic sense; rather, they impose different inductive biases on the same geometric input.
\\
The ablation results are consistent with the hypothesis that these representations are complementary, but stronger evidence would require unimodal and pairwise branch evaluations. Robustness claims also require controlled corruption experiments in which star detections, photometry, or image quality are systematically degraded.

\subsection{The discretization trade-off}
Replacing continuous angular regression with spherical classification removes the artificial RA wrap-around discontinuity and yields a simple coarse localization objective. The trade-off is quantization: with $K=15$, each class covers a large solid angle, so high classification accuracy does not imply fine pointing accuracy. Star-Fusion should therefore be interpreted as a coarse acquisition model rather than a final star-tracker attitude solution.

Standard cross entropy also treats all wrong classes equally, even though confusing a neighboring spherical cell is physically less severe than predicting a distant region. A natural extension is to use geometry-aware soft targets. For a ground-truth unit vector $\mathbf{v}$ and centroid $\mathbf{c}_k$, one possible target distribution is
\begin{equation}\label{eq:soft_labels}
q_k=\frac{\exp(\kappa\,\mathbf{v}^T\mathbf{c}_k)}{\sum_{j=1}^{K}\exp(\kappa\,\mathbf{v}^T\mathbf{c}_j)},
\end{equation}
where $\kappa$ controls the concentration around the true direction. Such a target would explicitly encode spherical proximity and could reduce boundary sensitivity.

\section{Limitations}
Several limitations must be addressed before Star-Fusion can be considered a complete flight-ready attitude solution:
\begin{itemize}
    \item \textbf{Boresight localization rather than full attitude:} The current target contains RA and Dec only and therefore does not recover rotation about the camera boresight. Full three-axis attitude estimation requires an additional roll degree of freedom or a direct representation of $SO(3)$, such as quaternions or rotation matrices.
    \item \textbf{Coarse quantization:} With $K=15$, the predicted regions are intentionally coarse. Classification accuracy should therefore be complemented by angular-error statistics and an accuracy--resolution study over multiple values of $K$.
    \item \textbf{Synthetic-domain evaluation:} The model is trained on synthetic Stellarium -derived imagery. Real sensors introduce non-uniform dark current, optical distortion, radiation-induced hot pixels, detector-specific point-spread functions, and other effects that may produce a domain gap.
    \item \textbf{No independent test set:} The present results are reported on a held-out validation split. Future evaluation should reserve a separate test set and report variability across multiple random seeds.
    \item \textbf{Unverified robustness to corrupted detections:} The present experiments do not yet isolate robustness to false stars, missed stars, centroid perturbations, magnitude noise, hot pixels, or strong motion blur.
    \item \textbf{High-angular-rate operation:} Motion blur at high spacecraft angular velocity can convert point-like stars into streaks. The present latency result establishes computational speed only; robustness under high-rate dynamics remains to be demonstrated empirically.
    \item \textbf{Coordinate-set representation:} A vanilla MLP is sensitive to the order of its input coordinates unless a deterministic convention is enforced. A permutation-invariant point-set encoder would provide a cleaner geometric formulation.
\end{itemize}

\section{Future Work}
Future work will extend the method from coarse boresight localization toward a complete LIS attitude pipeline. A natural direction is hierarchical coarse-to-fine estimation, where the current spherical classifier proposes one or more celestial regions and a second stage performs local angular or full-$SO(3)$ refinement. Geometry-aware soft labels, as in Eq.~\eqref{eq:soft_labels}, could reduce sensitivity to Voronoi boundaries.
\\
The coordinate branch can also be replaced by a permutation-invariant set encoder that processes each detected star with a shared feature map followed by symmetric pooling. Robustness should be evaluated systematically under centroid noise, missing stars, false detections, magnitude perturbations, hot pixels, and motion blur. A sweep over $K$ should quantify the trade-off between region resolution and classification accuracy, and the resulting models should be evaluated using angular-error statistics rather than Top-$k$ accuracy alone. Finally, temporal models across consecutive frames and real-sensor domain adaptation are promising directions for reducing ambiguity and closing the synthetic-to-flight gap.

\section{Conclusions}
This work introduced Star-Fusion, a three-branch architecture for coarse celestial boresight localization in Lost-in-Space conditions. By mapping RA/Dec directions to the unit sphere and predicting spherical cluster labels, the method avoids the artificial discontinuity of direct RA regression. The architecture combines photometric image features with rasterized and coordinate-based representations of detected star geometry.
\\
On a synthetic Stellarium -derived validation set with $K=15$, Star-Fusion achieved 93.4\% Top-1 accuracy, compared with 84.2\% for an image-only SwinV2-Tiny baseline. The measured neural forward-pass latency of 18.4~ms on an embedded NVIDIA Jetson platform indicates that the network itself is compatible with real-time embedded execution.
\\
These results support Star-Fusion as a coarse candidate-region estimator that can reduce the search space for a downstream fine attitude solver. They do not yet establish arc-second pointing accuracy, full three-axis attitude recovery, high-angular-rate robustness, or end-to-end flight readiness. Addressing these points through angular-error evaluation, stronger ablations, corruption tests, and real-sensor validation is the main direction for future work.

\section*{Acknowledgements}
The authors thank the Department of Aerospace Informatics at Julius-Maximilians-Universit\"at W\"urzburg for supporting this research.


\begin{thebibliography}{99}
\bibitem{mortari1997pyramid} D. Mortari, The Pyramid star identification technique, \textit{Navigation, Journal of the Institute of Navigation}, 44(3) (1997) 345--358.
\bibitem{padgett1997grid} J. L. Padgett and K. Kreutz-Delgado, A grid algorithm for star identification, \textit{IEEE Transactions on Aerospace and Electronic Systems}, 33(1) (1997) 202--213.
\bibitem{liu2022swinv2} Z. Liu et al., Swin Transformer V2: Scaling up capacity and resolution, \textit{Proceedings of the IEEE/CVF Conference on Computer Vision and Pattern Recognition (CVPR)}, 2022, pp. 12009--12019.
\bibitem{dosovitskiy2020vit} A. Dosovitskiy et al., An image is worth 16x16 words: Transformers for image recognition at scale, \textit{arXiv preprint arXiv:2010.11929}, 2020.
\bibitem{cohen2018spherical} T. S. Cohen, M. Geiger, J. K\"ohler, and M. Welling, Spherical CNNs, \textit{International Conference on Learning Representations (ICLR)}, 2018.
\bibitem{arthur2007kmeanspp} D. Arthur and S. Vassilvitskii, k-means++: The advantages of careful seeding, \textit{Proceedings of the Eighteenth Annual ACM-SIAM Symposium on Discrete Algorithms (SODA)}, 2007, pp. 1027--1035.
\bibitem{rijlaarsdam2020efficient} D. Rijlaarsdam, H. Yous, J. Byrne, D. Oddenino, D. Furano, and D. Moloney, Efficient star identification using a neural network, \textit{Sensors}, 20(13) (2020) 3684.
\bibitem{wang2020ai} H. Wang, Z. Wang, B. Wang, Z. Yu, Z. Jin, and J. L. Crassidis, An artificial intelligence enhanced star identification algorithm, \textit{Frontiers of Information Technology \& Electronic Engineering}, 21(11) (2020) 1661--1670.
\bibitem{wang2021cnn} B. Wang, H. Wang, and Z. Jin, An efficient and robust star identification algorithm based on neural networks, \textit{Sensors}, 21(22) (2021) 7686.
\end{thebibliography}
\end{document}